\documentclass[letterpaper]{article} 
\usepackage{aaai2026}  
\usepackage{times}  
\usepackage{helvet}  
\usepackage{courier}  
\usepackage[hyphens]{url}  
\usepackage{graphicx} 
\usepackage{natbib}  
\usepackage{caption} 
\usepackage{algorithm}
\usepackage{algorithmic}

\usepackage{amsmath}
\usepackage{booktabs}

\usepackage{newfloat}
\usepackage{listings}
\DeclareCaptionStyle{ruled}{labelfont=normalfont,labelsep=colon,strut=off} 
\floatstyle{ruled}
\newfloat{listing}{tb}{lst}{}
\floatname{listing}{Listing}
\title{Pruning Large Language Models by Identifying and Preserving Functional Networks}
\author{
    Yiheng Liu\textsuperscript{\rm 1}, Junhao Ning\textsuperscript{\rm 1}, Sichen Xia\textsuperscript{\rm 1}, Xiaohui Gao\textsuperscript{\rm 1}, Ning Qiang\textsuperscript{\rm 2}, Bao Ge\textsuperscript{\rm 2}, Junwei Han\textsuperscript{\rm 1}, Xintao Hu\textsuperscript{\rm 1}
}
\affiliations{
    \textsuperscript{\rm 1}School of Automation, Northwestern Polytechnical University, Xi'an, China\\
    \textsuperscript{\rm 2}School of Pyhisic and Information Technology, Shaanxi Normal University, Xi'an
}

\usepackage{bibentry}

\begin{document}

\maketitle

\begin{abstract}
Structured pruning is one of the representative techniques for compressing large language models (LLMs) to reduce GPU memory consumption and accelerate inference speed. It offers significant practical value in improving the efficiency of LLMs in real-world applications. Current structured pruning methods typically rely on assessment of the importance of the structure units and pruning the units with less importance. Most of them overlooks the interaction and collaboration among artificial neurons that are crucial for the functionalities of LLMs, leading to a disruption in the macro functional architecture of LLMs and consequently a pruning performance degradation. Inspired by the inherent similarities between artificial neural networks and functional neural networks in the human brain, we alleviate this challenge and propose to prune LLMs by identifying and preserving functional networks within LLMs in this study. To achieve this, we treat an LLM as a digital brain and decompose the LLM into functional networks, analogous to  identifying functional brain networks in neuroimaging data. Afterwards, an LLM is pruned by preserving the key neurons within these functional networks. Experimental results demonstrate that the proposed method can successfully identify and locate functional networks and key neurons in LLMs, enabling efficient model pruning. Our code is available at \url{https://github.com/WhatAboutMyStar/LLM_ACTIVATION}.
\end{abstract}


\section{Introduction}
Modern large language models (LLMs) often contain tens of billions to hundreds of billions of parameters, resulting in significant computational costs during both training and deployment \cite{zhao2023survey,liu2023summary,WANG2024,LIU2025129190}. Among various model compression approaches, pruning has gained attention as an effective approach to address this challenge. By removing redundant parameters while maintaining model performance, pruning not only substantially reduces GPU memory usage, but also accelerates inference \cite{cheng2024survey}.  

Model pruning methods can be broadly categorized into structured and unstructured pruning, distinguished by their respective targets of structural components versus individual weights \cite{ma2023llm,frantar2023sparsegpt,sun2023simple,bai2024sparsellm}. While unstructured pruning sparsifies the weight matrix by setting redundant weights to zero, structured pruning removes architectural elements such as attention heads, neuron groups, or layers in a physically meaningful manner \cite{ma2023llm,men2024shortgpt,kim2024shortened,kim2024mefomo,gromov2025the}. It inherently improves inference speed and memory efficiency through actual model size reduction, while maintaining compatibility with standard hardware architectures.

Structured pruning typically evaluates and removes less important structural units within LLMs. Candidate structural units include individual neurons \cite{an2024fluctuation,ashkboos2024slicegpt}, a group of coupled structures identified based on pairwise neuron dependencies \cite{ma2023llm}, and entire Transformer blocks \cite{men2024shortgpt, song2024sleb,kim2024shortened,chenstreamlining}, etc. Although these methods have shown promise in reducing model size, they largely overlook the integrity of the functional cohesion of neural structures in LLMs. This stands in contrast to biological neural systems, where cognition emerges from coordinated neural activities that result from structural connections and functional interactions between neurons. This principle, known as functional brain networks, is well-established in neuroscience \cite{bullmore2009complex}.

Inspired by the inherent similarities between artificial neural networks and functional networks in the human brain \cite{liu2025brain}, we propose a novel structured pruning method that identifies and preserves functional networks within LLMs to minimize architectural disruption during the pruning process. To this end, we conceptualize an LLM as a digital brain where individual artificial neurons are viewed as the fundamental processing units. We analyze LLMs in a similar manner to how fMRI data is analyzed, treating the neurons signals of LLMs as voxels in fMRI signals. We then applied independent component analysis, which is a well-established method to identify functional brain networks in neuroimaging data, to discover coherent functional networks with LLMs. By identifying and preserving as many of these functional networks as possible, we can achieve structured pruning of LLMs. The proposed method achieves state-of-the-art performance on multiple benchmark datasets, validating the effectiveness of our brain-inspired approach.

Our contributions can be summarized as follows:

1. We propose an effective structured pruning method that significantly reduces the computational and memory requirements of LLMs while preserving their performance as close to the original as possible.
 
2. We introduce analysis method from cognitive neuroscience into the study of identifying functional networks within LLMs, demonstrating their effectiveness in understanding and improving these models. This work provide preliminary evidences for understanding the functional mechanism of LLMs and for designing more biologically plausible and interpretable LLMs.

\section{Related Works}
Model pruning is a model compression technique \cite{lecun1989optimal,han2015learning,hassibi1993optimal} that can be divided into structured pruning and unstructured pruning. Unstructured pruning operates on individual weights within the parameter matrix. Although it can achieve significant sparsity, the weights are only set to zero in a logical sense, which does not effectively improve the inference speed or reduce GPU memory consumption \cite{han2016deep,zafrir2021prune}. A special variant of unstructured pruning, known as semi-structured pruning, requires that exactly $N$ non-zero values must be preserved in every block of $M$ consecutive weights \cite{zhou2021learning}. In general, these unstructured pruning methods rely on hardware acceleration for sparse matrices to deliver performance improvements. In contrast, structured pruning physically removes individual neurons or entire structural units, offering a solution that is more aligned with hardware characteristics \cite{xia2022structured,molchanov2017pruning}. This approach not only reduces GPU memory consumption but also significantly enhances inference speed. In this section, we will focus on the related works of structured pruning for LLMs. 

Based on the specific aspects of the model structure that is being pruned, such as layers, blocks, or neurons, the model pruning methods can be classified as depth pruning and width pruning. Recently, depth pruning for LLMs has gained significant attention, with several studies focusing on this area. These approaches primarily aim to define metrics that measure the importance of layers, followed by removing redundant Transformer blocks within LLMs to achieve efficient model compression. Notable works in this category include ShortGPT \cite{men2024shortgpt}, SLEB \cite{song2024sleb}, Shortened LLaMA \cite{kim2024shortened}, and LLM-Streamline \cite{chenstreamlining}. However, directly removing entire redundant Transformer blocks is too coarse-grained and may disrupt the functional architecture present in LLMs. Although this approach is more suitable for accelerating the inference speed, it remains insufficient to preserve the performance of the model.

Our work falls into the category of width pruning. A key aspect of width pruning is to evaluate the importance of artificial neurons (hidden dimensions) or structural components in the model and remove those that are deemed less important. For example, FLAP \cite{an2024fluctuation} evaluates the importance of neurons by measuring the fluctuation of each channel across different layers and modules. LLM-Pruner \cite{ma2023llm} identifies groups of coupled structure within LLMs based on the dependency of all possible pair of neurons and employs an importance estimation strategy to select the optimal group for pruning. Although LLM-Pruner aims to minimize architectural disruption during pruning, its reliance on simplified pairwise neuron dependencies presents notable limitations. In addition to evaluation-based pruning methods, recent work has also explored a data-driven approach for structured pruning. For example, SliceGPT \cite{ashkboos2024slicegpt} utilizes Principal Component Analysis (PCA) to remove rows or columns with low distinguishability in the parameter matrix. However, PCA selects embedding directions based solely on variance, rather than functional relevance. 

In summary, existing width pruning methods often overlook the integrity of functional cohesion among neural structures in LLMs. Pruning is usually done based only on the importance of structural units, which may ignore groups of neurons that work collectively. As a result, this can lead to less effective pruning performance.

\section{Preliminaries}
In this section, we introduce some background knowledge that is necessary to understand the content of this study, including which structural units in LLMs under the Transformer architecture are pruned, as well as how neuroscience investigates functional networks in the human brain. Furthermore, it explores how the human brain and artificial neural networks like LLMs can be aligned to analyze LLMs using neuroscientific methods.

\textbf{Transformer, MLP:} The Multi-Layer Perceptron (MLP) layer plays a crucial role in introducing non-linearity and enhancing the expressive power of models in the Transformer architecture. In the context of LLMs, this layer is also referred to as the Position-wise Feed-Forward Network (FFN). The FFN applies two linear transformations. Typically, the hidden dimension $d_{\text{hidden}}$ is set to be slightly larger than the original dimension $d_{\text{model}}$ of the model.. For example, in models such as GPT-2, the hidden dimension is expanded to $4 \cdot d_{\text{model}}$. However, this expansion factor is not strictly fixed. In the LLaMA series \cite{touvron2023llama}, the input dimension is expanded from 4096 to 11008, the exact expansion ratio depending on the design of the model. Following the first linear transformation, a non-linear activation function is applied to introduce non-linearity. The second linear transformation projects the output back to the original dimension $d_{\text{model}}$.

\begin{figure}
    \centering
    \includegraphics[width=\linewidth]{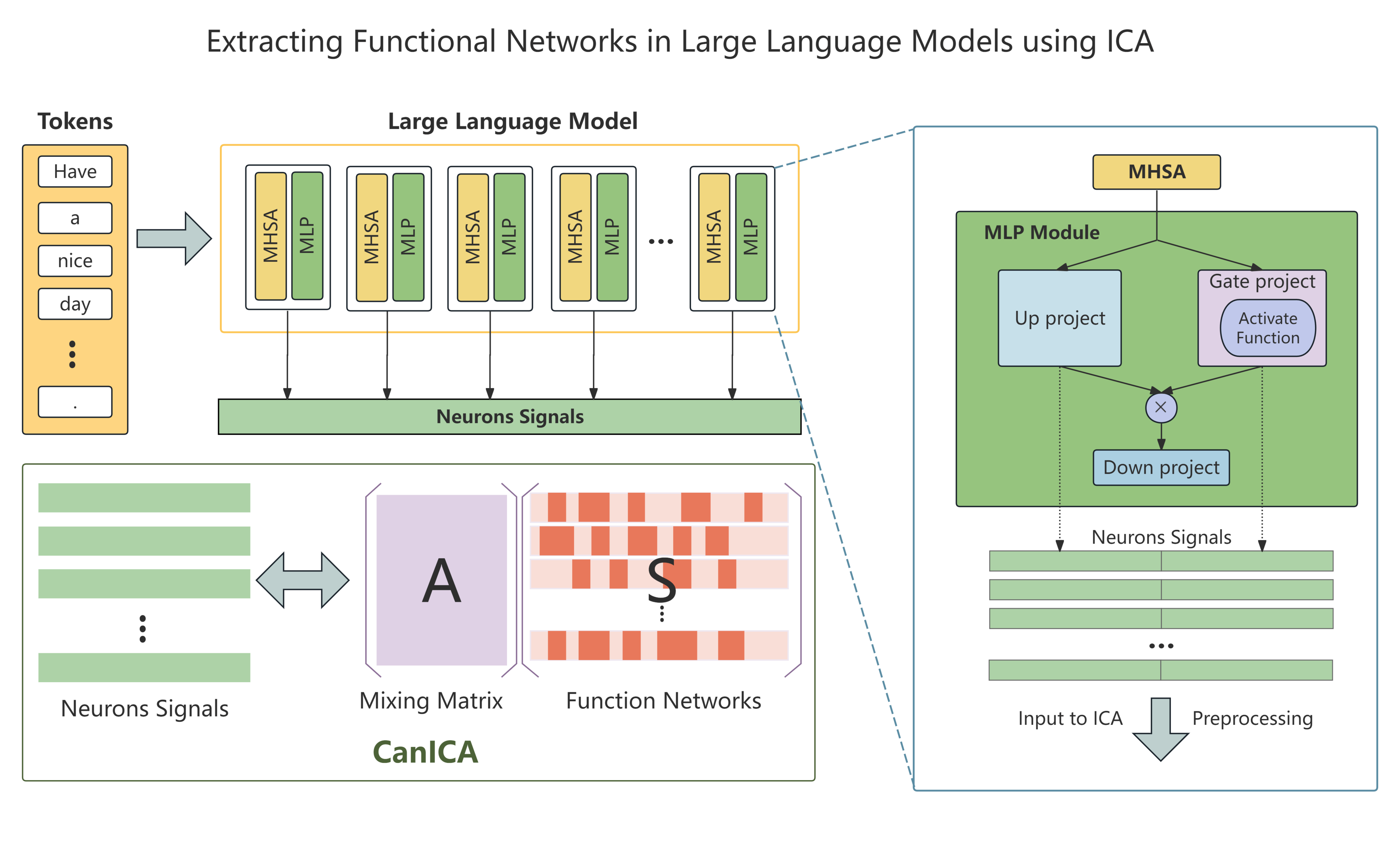}
    \caption{The pipeline of identifying functional networks within LLMs. The temporal signals of MLP neurons are aggregated and an spatial ICA algorithm canonical ICA (CanICA) is applied to the signal matrix to identify functional networks in LLMs.}
    \label{fig:llm-ica}
\end{figure}

It is worth noting that in modern LLMs, bias terms are often omitted in both linear transformations. In modern LLMs, a gating mechanism is typically introduced within the MLP layer. This involves adding an additional linear layer, where the gate projection maps the output through an activation function before multiplying it by the linear layer of the up projection as shown in Figure~\ref{fig:llm-ica}. This design allows for more nuanced control over information flow, enhancing the model's capacity to capture complex patterns and improve overall performance. In this paper, we will focus on pruning the MLP layers. 

\textbf{Functional Brain Networks, FBN:} A functional brain network is defined as a collection of brain regions that exhibit coordinated activation during cognitive tasks or at resting state. These networks can be indirectly observed using functional Magnetic Resonance Imaging (fMRI), which measures Blood-Oxygen Level Dependent (BOLD) signals to infer neuronal activity. In fMRI data, the intensity of the signal of each voxel corresponds to the BOLD response of a localized brain region at a specific time point. 
 
Importantly, fMRI signals do not represent direct recordings of individual neural processes. Instead, they reflect the linear superposition of multiple independent functional brain networks, such as default mode network (DMN) \cite{raichle2015brain}, dorsal attention network (DAN) \cite{szczepanski2013functional}, and others \cite{smith2009correspondence}. This implies that the interpretation of fMRI data requires accounting for the complex interactions among these functional brain networks. 

The identification and characterization of functional brain networks have become pivotal techniques in neuroscience research. By uncovering the organization and dynamics of functional brain networks, researchers can gain deeper insight into the neural mechanisms underlying cognition, ultimately advancing our understanding of both normal brain function and neurological or psychiatric disorders. In this study, we hypothesize that LLMs possess analogous functional networks. Identifying these functional networks is essential for the effective pruning of LLMs.

\begin{figure}
    \centering
    \includegraphics[width=\linewidth]{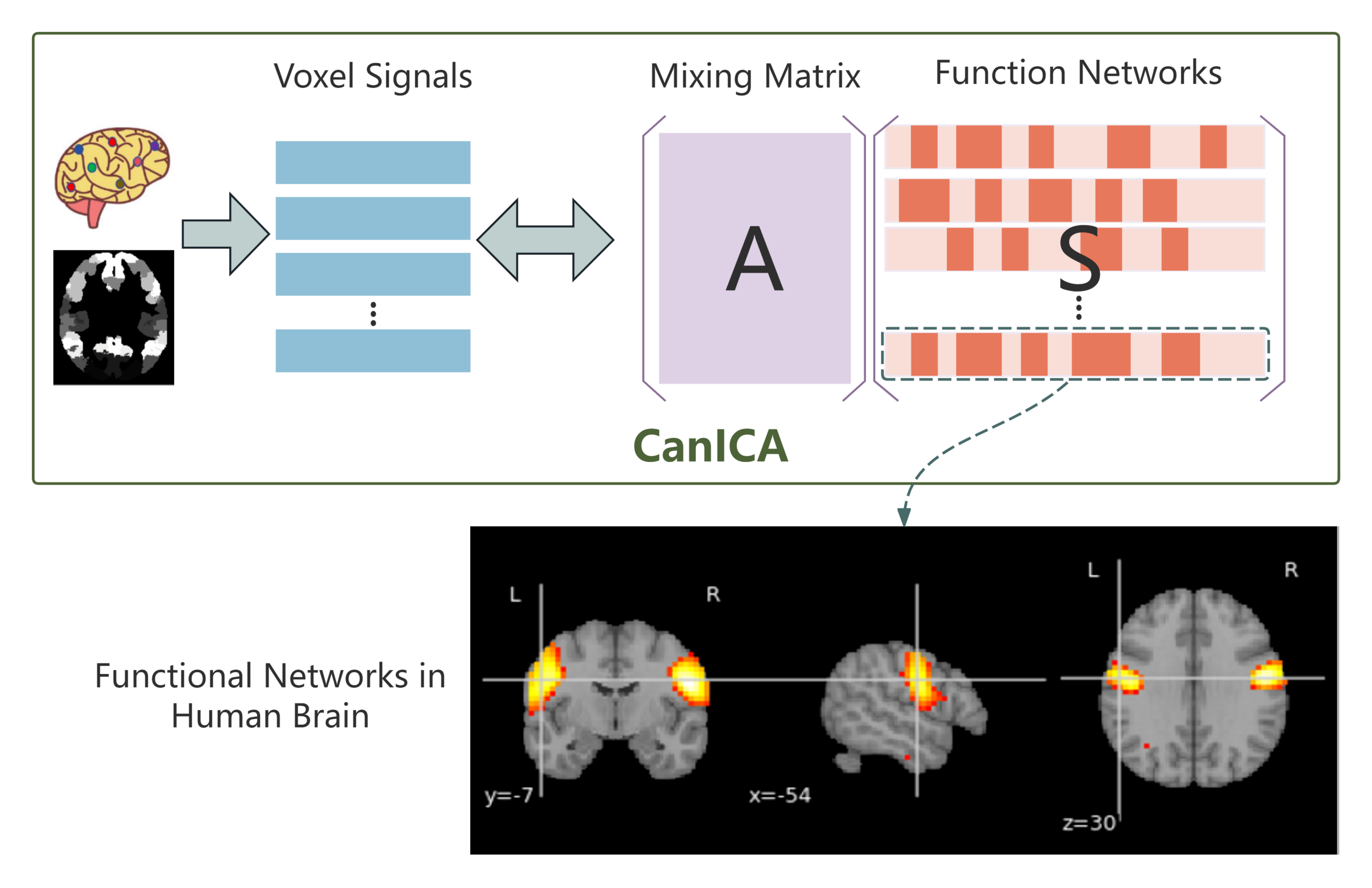}
    \caption{The pipline of extracting functional networks in fMRI signals using CanICA.}
    \label{fig:fmri-ica}
\end{figure}

\textbf{Independent Component Analysis, ICA:} ICA is a data-driven approach that is widely adopted to identify functional brain networks from high-dimensional fMRI data. ICA assumes that the observed BOLD signals are linear combinations of several non-Gaussian, statistically independent source signals. The spatial patterns of these independent components depict the spatial distribution of functional brain networks, as illustrated in Figure~\ref{fig:fmri-ica}. 
 
The successful application of ICA in fMRI analysis provides a valuable methodological framework for identifying functional networks within LLMs. There are notable parallels between fMRI signals and neuron signals in LLMs. While fMRI signals reflect the activity of biological neurons in response to various stimuli, the output signals of neurons in LLMs represent the internal state changes as the model processes input information. Both represent forms of "neural activity", but they belong to different areas: fMRI is related to biological brains, while LLMs are related to artificial neural networks. Additionally, both fMRI and LLM neuron signals exhibit temporal dynamics. fMRI records activity changes in different brain regions over time in response to stimuli, whereas LLM neurons' outputs vary at different time steps depending on input data. By interpreting neuronal signals in LLMs (e.g., intermediate feature vectors from Transformer layers) as analogous to "BOLD signals," the ICA framework can be adapted to extract independent functional components within LLMs. 

ICA reconstructs latent source signals $\mathbf{S}$ from the observed mixed signals $\mathbf{X}$. Assume we have $n$ observed signals $\left[ \mathbf{x}_1, \mathbf{x}_2, \ldots, \mathbf{x}_n \right]$, which are linear combinations of $m$ statistically independent source signals $\left[ \mathbf{s}_1, \mathbf{s}_2, \ldots, \mathbf{s}_m \right]$. The relationship between the observed signals $\mathbf{X}$ and the source signals $\mathbf{S}$ can be mathematically expressed as: $\mathbf{X} = \mathbf{A} \mathbf{S}$. Where $\mathbf{A}$ denotes the mixing matrix that describes how the source signals are linearly combined to generate the observed signals. Each column of the mixing matrix $\mathbf{A}$ corresponds to the spatial pattern of a specific functional network, indicating which brain regions exhibit coordinated activation. In the context of this study, the functional networks derived from the neuronal signals of LLMs correspond to the rows of the source signal matrix $\mathbf{S}$. In this study, we adopt canonical ICA (CanICA) algorithm for performing ICA analysis \cite{varoquaux2010group,varoquaux2010ica}. CanICA is a spatial ICA method that differs from traditional ICA algorithms. When using conventional ICA to extract functional networks, the functional networks would correspond to the mixing matrix $\mathbf{A}$. However, in spatial ICA like CanICA, the functional networks correspond to the source signal matrix $\mathbf{S}$ instead.

\section{Method}
\subsection{Datasets and Models}
Consistent with most existing approaches, we use the Wikitext-2 dataset \cite{merity2016pointer} as the calibration dataset for pruning LLMs. We derive neuron signals from the outputs of the Gate Projection and the Up Projection in the MLP module as shown in Figure~\ref{fig:llm-ica}. These neuronal signals are preprocessed by z-score standardization, similar to the preprocessing of fMRI data in neuroscience. This involves applying z-score standardization individually to each neuron, which helps the ICA focus more effectively on the functional activities of neurons that are activated simultaneously. The preprocessed neural signals are then inputted into the CanICA algorithm, as illustrated in Figure~\ref{fig:llm-ica}. 

\subsection{Identify and Preserve Functional Networks}
\begin{figure}
    \centering
    \includegraphics[width=\linewidth]{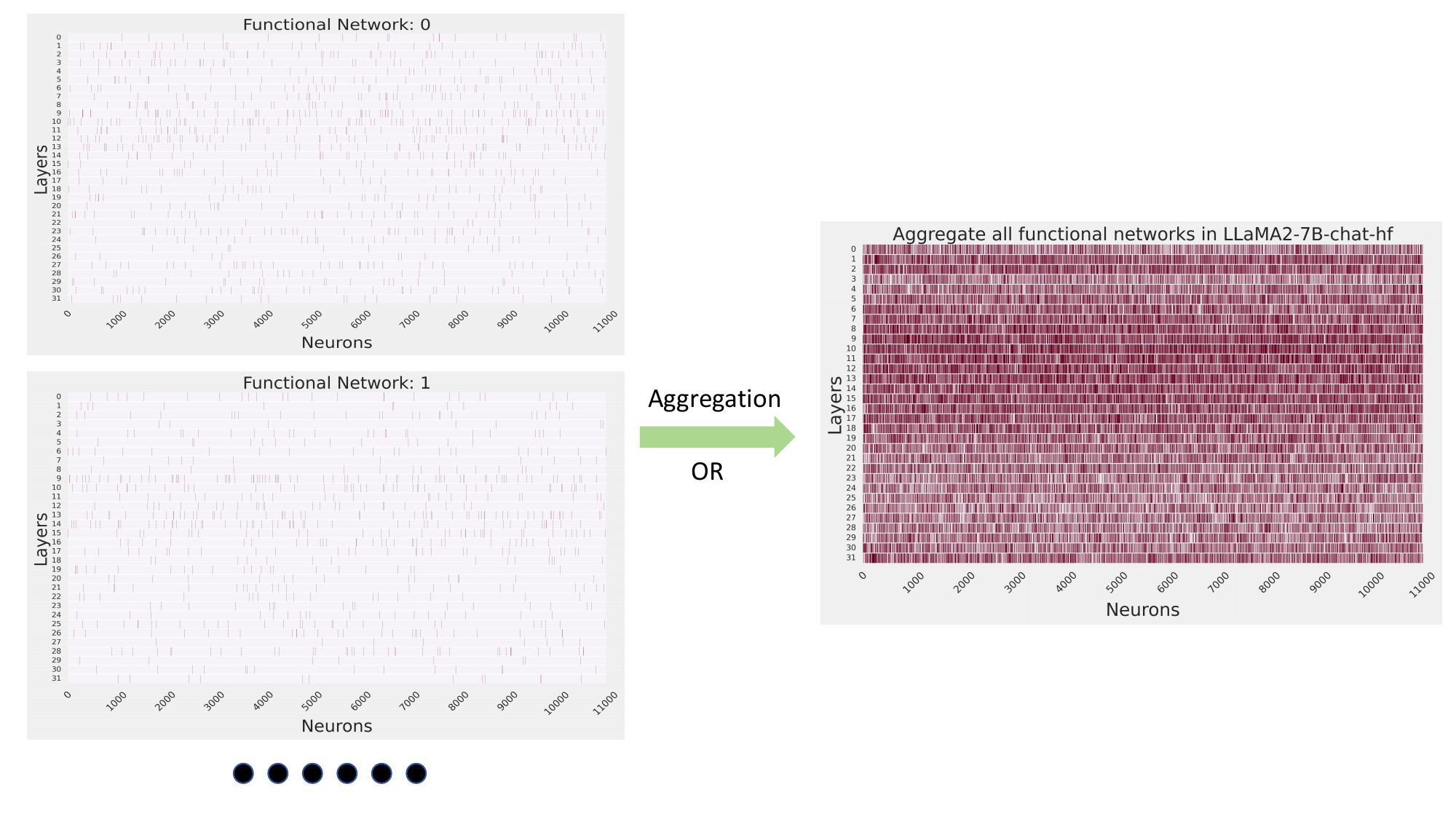}
    \caption{The process of aggregating the mask matrices of all obtained functional networks through an OR operation to produce the final global mask matrix used for pruning.}
    \label{fig:fbn}
\end{figure}

The vast number of artificial neurons in LLMs and the high dimensionality of their signals pose significant challenges to analyzing LLMs using ICA, which are commonly employed for studying human brains. The number of neurons in LLMs far exceeds the number of voxels in fMRI data, leading to a substantial increase in computational complexity. This not only requires more memory resources but also makes the analysis less likely to converge efficiently, further complicating the process of extracting meaningful insights from the neural signals of LLMs. Traditional ICA is used to analyze functional networks in the human brain and typically processes voxel data within the range of no more than 100,000. In contrast, for LLMs, the enormous number of neurons results in ICA having to handle extremely large datasets, leading to slow computational speeds and making it difficult to meet practical research requirements. For example, in LLM models with scales of 6B or 7B parameters, the total number of neurons can easily reach millions. Not to mention the even larger numbers of neurons in 9B, 13B, 30B, 70B, 130B, 230B, 631B or larger-scale LLMs. 

To overcome this, we propose a layer-by-layer ICA-based local pruning method, which performs CanICA independently on the neuron signals of each layer of the LLM. After applying CanICA, we obtain the functional networks corresponding to each layer as shown in Figure~\ref{fig:llm-ica}. Notably, CanICA is a form of spatial ICA that assumes the independence of source signals in the spatial domain rather than the temporal domain. This approach not only facilitates the capture of distinct functional networks within the high-dimensional outputs of neurons but also effectively reduces the dimensionality of the neuron signals and decreases computational complexity, making it advantageous for analyzing large-scale neural data in high-dimensional spaces.

Identified functional networks are represented by the source signal matrix, which consists of $k$ rows. Each row corresponds to a functional network, and each column represents the associated neurons. By applying thresholding to the source signal matrix, we can obtain the mask matrices of the functional networks, which are used to identify the critical neurons that need to be preserved during pruning.

Since the association of neurons with specific functional networks can vary depending on different calibration data samples, to ensure effective pruning, it is necessary to extract as many functional networks as possible, thereby ensuring that all important neurons are retained.

In our experiments, we divided 3200 calibration data samples into multiple groups and executed CanICA multiple times. For each run, we input 40 samples and extract 128 functional networks. Finally, we aggregated mask matrices from different groups using an OR operation to generate a global mask matrix as shown in Figure~\ref{fig:fbn}, which was then applied for pruning.

\begin{table*}
	\caption{The results of 20\% pruning rate for Vicuna-7B-v1.5. The results marked in bold represent the best performance.}\label{tab:vicuna20}
	\centering
	\begin{tabular}{ccccccccc}
    \toprule
    Pruning Rate & Method & PIQA & WinoGrande & HellaSwag & ARC-E & ARC-C & OBQA & Wikitext2 \\ 
    \midrule
    0\%          & Original Model                       & 0.7726 & 0.6953     & 0.5645    & 0.7563   & 0.4317        & 0.3300 & 11.3191   \\ 
    20\%         & LLM-Pruner                           & \textbf{0.7361} & 0.5714     & 0.4789    & 0.6810   & 0.3754        & 0.2800 & 27.3497   \\ 
    20\%         & SliceGPT                             & 0.6425 & 0.5967     & 0.3753    & 0.5568   & 0.2910        & 0.2100 & 60.8797   \\ 
    20\%         & FLAP                                 & 0.7291 & 0.6630     & 0.4963    & 0.6848   & 0.3498        & 0.2760 & 16.2735   \\ 
    20\%         & FLAP (without bias)                  & 0.7225 & 0.6575     & 0.4913    & 0.6793   & 0.3541        & 0.2620 & 16.6949   \\ 
    20\%         & Shortened LLaMA                      & 0.7078 & 0.6535     & 0.4790    & 0.6368   & \textbf{0.3805}        & 0.2640 & 25.8674   \\
    20\%         & CanICA & 0.7301 & 0.6669 & \textbf{0.4974} & 0.6839 & 0.3669 & \textbf{0.306} & 16.3667 \\
    20\%         & CanICA (add bias)  & 0.7259 & \textbf{0.6772} & 0.4950 & \textbf{0.6919} & 0.3737 & 0.296 & \textbf{15.8466} \\
    \bottomrule
\end{tabular}
\end{table*}

\begin{table*}
	\caption{The results of 20\% pruning rate for LLaMA2-7B-chat-hf. The results marked in bold represent the best performance.}\label{tab:llama20}
	\centering
	\begin{tabular}{ccccccccc}
    \toprule 
    Pruning Rate & Method & PIQA & WinoGrande & HellaSwag & ARC-E & ARC-C & OBQA & Wikitext2 \\ 
    \midrule 
    0\%          & Original Model                       & 0.7639 & 0.6646     & 0.5779    & 0.7386   & 0.4420        & 0.3320 & 11.5959   \\ 
    20\%         & LLM-Pruner                           & 0.7176 & 0.5738     & 0.4826    & 0.6561   & 0.3592        & 0.2700 & 28.6396   \\ 
    20\%         & SliceGPT                             & 0.6328 & 0.5998     & 0.3724    & 0.5501   & 0.2901        & 0.2160 & 72.2155   \\ 
    20\%         & FLAP                                 & 0.7356 & 0.6496     & \textbf{0.5051}    & 0.6612   & 0.3515        & \textbf{0.3140} & 17.0853   \\ 
    20\%         & FLAP (without bias)                  & \textbf{0.7383} & 0.6361     & 0.4987    & 0.6477   & 0.3439        & 0.3120 & 18.2824   \\ 
    20\%         & Shortened LLaMA                      & 0.7155 & 0.6440     & 0.4889    & 0.6233   & \textbf{0.3857}        & 0.2600 & 28.9063   \\
    20\%         & CanICA                               & 0.7220 & 0.6527     & 0.5044    & 0.6620   & 0.3532        & 0.2800 & 17.9802   \\ 
    20\%         & CanICA (add bias)                    & 0.7214 & \textbf{0.6598}     & 0.4997    & \textbf{0.6713}   & 0.3814        & 0.2900 & \textbf{16.6894}   \\ 
    \bottomrule
\end{tabular}
\end{table*}

\begin{table*}
	\caption{The results of different pruning rates for ChatGLM3-6B-base.}\label{tab:chatglm}
	\centering 
	\begin{tabular}{ccccccccc}
    \toprule 
    Pruning Rate & Method & PIQA & WinoGrande & HellaSwag & ARC-E & ARC-C & OBQA & Wikitext2 \\ 
    \midrule 
    0\%          & Original Model                       & 0.7900 & 0.7253     & 0.5929    & 0.7992   & 0.4966        & 0.3120 & 10.1177   \\ 
    10\%         & CanICA & 0.7829 & 0.7040     & 0.5663    & 0.7753   & 0.4684        & 0.3020 & 11.4666   \\ 
    20\%         & CanICA & 0.7590 & 0.6946     & 0.5308    & 0.7273   & 0.4292        & 0.2720 & 13.9871   \\ 
    30\%         & CanICA & 0.7247 & 0.6693     & 0.4621    & 0.6587   & 0.3746        & 0.2540 & 20.3200   \\ 
    \bottomrule
\end{tabular}
\end{table*}

\begin{table*}
	\caption{The results of 30\% pruning rate for Vicuna-7B-v1.5. The results marked in bold represent the best performance.}\label{tab:vicuna30}
	\centering
	\begin{tabular}{ccccccccc}
    \toprule 
    Pruning Rate & Method & PIQA & WinoGrande & HellaSwag & ARC-E & ARC-C & OBQA & Wikitext2 \\ 
    \midrule 
    0\%          & Original Model                       & 0.7726 & 0.6953     & 0.5645    & 0.7563   & 0.4317        & 0.3300 & 11.3191   \\ 
    30\%         & LLM-Pruner                           & 0.6975 & 0.5343     & 0.4387    & 0.6216   & \textbf{0.3387}        & \textbf{0.2620} & 42.8910   \\ 
    30\%         & SliceGPT                             & 0.5653 & 0.5083     & 0.3060    & 0.3742   & 0.2201        & 0.1600 & 177.1559   \\ 
    30\%         & FLAP                                 & 0.6991 & 0.6314     & \textbf{0.4537}    & \textbf{0.6490}   & 0.3259        & \textbf{0.2620} & \textbf{21.4575}   \\ 
    30\%         & FLAP (without bias)                  & 0.6964 & 0.6188     & 0.4503    & 0.6292   & 0.3362        & \textbf{0.2620} & 24.7562   \\ 
    30\%         & Shortened LLaMA                      & 0.6181 & 0.6488     & 0.3789    & 0.4731   & 0.3183        & 0.1920 & 244.8821   \\
    30\%         & CanICA                               & \textbf{0.7062} & \textbf{0.6504}     & 0.4369    & 0.6195   & 0.3276        & 0.2580 & 25.9746   \\ 
    30\%         & CanICA (add bias)                    & 0.6975 & 0.6306     & 0.4244    & 0.6199  & 0.3217      & 0.2460 & 24.1499  \\ 
    \bottomrule
\end{tabular}
\end{table*}

\begin{figure*}
    \centering
    \includegraphics[width=\linewidth]{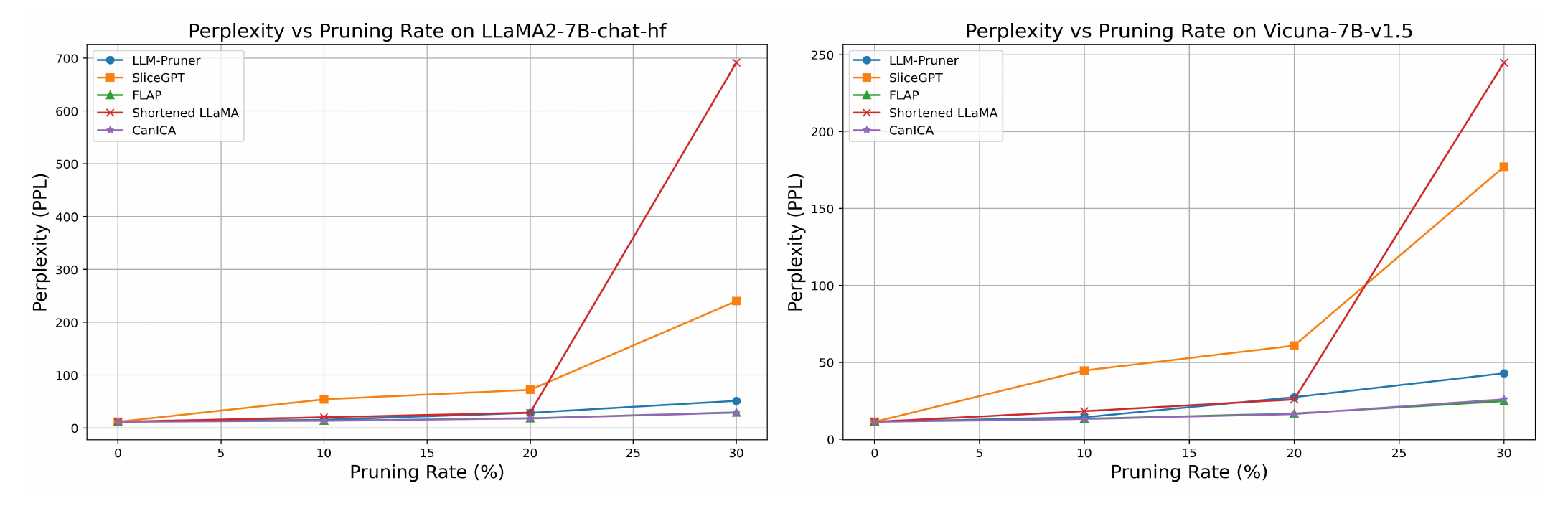}
    \caption{The perplexity results on LLaMA2-7B-chat-hf (left) and Vicuna-7B-v1.5 (right) with different pruning rates.}
    \label{fig:ppl}
\end{figure*}

\begin{table*}
	\caption{The results of 30\% pruning rate for LLaMA2-7B-chat-hf. The results marked in bold represent the best performance.}\label{tab:llama30}
	\centering
	\begin{tabular}{ccccccccc}
    \toprule 
    Pruning Rate & Method & PIQA & WinoGrande & HellaSwag & ARC-E & ARC-C & OBQA & Wikitext2 \\ 
    \midrule 
    0\%          & Original Model                       & 0.7639 & 0.6646     & 0.5779    & 0.7386   & 0.4420        & 0.3320 & 11.5959   \\ 
    30\%         & LLM-Pruner                           & 0.6904 & 0.5525     & 0.4248    & 0.5589   & 0.3038        & 0.2320 & 51.3416   \\ 
    30\%         & SliceGPT                             & 0.5707 & 0.5257     & 0.3055    & 0.3868   & 0.2193        & 0.1540 & 240.0398   \\ 
    30\%         & FLAP                                 & 0.7013 & 0.6314     & \textbf{0.4519}    & \textbf{0.6423}   & \textbf{0.3515}        & \textbf{0.2680} & \textbf{23.9317}   \\ 
    30\%         & FLAP (without bias)                  & \textbf{0.7029} & 0.6125     & 0.4449    & 0.6317   & 0.3473        & 0.2600 & 29.5235   \\ 
    30\%         & Shortened LLaMA                      & 0.6159 & 0.6212     & 0.3653    & 0.4444   & 0.3072        & 0.1860 & 691.5356   \\
    30\%         & CanICA                               & 0.6986 & 0.6243     & 0.4366    & 0.5960   & 0.3268        & 0.2460 & 29.0336   \\ 
    30\%         & CanICA (add bias)                    & 0.6980 & \textbf{0.6330}     & 0.4312    & 0.6199  & 0.3251       & 0.2540 & 25.4853  \\ 
    \bottomrule
\end{tabular}
\end{table*}

\section{Results}
\subsection{Experimental Settings}
To facilitate direct comparisons with other methods, we conduct pruning experiments on LLaMA2-7B-chat-hf \cite{touvron2023llama} and Vicuna-7B-v1.5 \cite{chiang2023vicuna} models. For evaluation, we employ the lm-evaluation-harness framework \cite{eval-harness} to assess the performance of the pruned model. Specifically, we evaluate the perplexity of the pruned model in the Wikitext-2 dataset \cite{merity2016pointer} and its zero-shot classification performance on several benchmark datasets, including PIQA \cite{bisk2020piqa}, HellaSwag \cite{zellers2019hellaswag}, Winogrande \cite{sakaguchi2021winogrande}, OpenBookQA \cite{mihaylov2018can}, ARC-Easy \cite{clark2018think}, and ARC-Challenge \cite{clark2018think}. Additionally, we also evaluate the pruning performance of the proposed method on ChatGLM3-6B-base \cite{glm2024chatglm} model to validate the proposed method on models with different architectures. 

The proposed method is compared to representative up-to-date structured pruning methods, including LLM-Pruner \cite{ma2023llm}, FLAP \cite{an2024fluctuation}, SliceGPT \cite{ashkboos2024slicegpt}, and Shortened LLaMA \cite{kim2024shortened,kim2024mefomo}. Notably, FLAP employs a compensation bias term after pruning. For fair comparisons, we provide both the results of FLAP without the compensation bias term and the results of the proposed method with the identical compensation bias term. To ensure a fair comparison, all pruning methods did not use recovery training to restore the model's performance.

The number of independent components in CanICA is empirically set as 128. We also conduct an experiment to evaluate the impact of different numbers of independent components on the performance of the pruned model.

\subsection{Comparison Results}
Structured pruning methods usually compare the performance of pruning 20\% of the original model. We present the results after pruning 20\% of the Vicuna-7B-v1.5 and LLaMA2-7B models in Table~\ref{tab:vicuna20} and Table~\ref{tab:llama20}, respectively. 

In Vicuna-7B-v1.5, the proposed method (CanICA, the 7-th row in Table~\ref{tab:vicuna20}), achieves the best performance in 5 out of the in total 7 tasks when compared to LLM-Pruner, SliceGPT, FLAP without bias compensation and Shortened LLaMA. When accounting for the compensation bias term, the proposed method (CanICA add bias, the 8-th row in Table~\ref{tab:vicuna20}) outperforms the standard FLAP approach in 5 out of the in total 7 tasks. The highest performance gain (2.39\%) is achieved in the ARC-C task. CanICA also decreases the perplexity metric in Wikitext2 task by 0.3269. In LLaMA2-7B, the proposed method achieves the best performance in 4 tasks (notably FLAP without bias compensation for comparison). When bias compensation is used, the proposed method outperforms FLAP in 4 out of the 7 tasks. 

We also evaluate the proposed method on ChatGLM3-6B-base. The network architecture of ChatGLM3-6B-base slightly differs from the LLaMA series, primarily in the up projection of the MLP module. Specifically, ChatGLM3-6B-base uses a larger expansion dimension in this component compared to the LLaMA models. Since other methods do not currently support pruning for ChatGLM3-6B-base, the pruning results for this model are solely based on our method. The pruning results are shown in Table~\ref{tab:chatglm}. It is seen that the proposed method achieves the best performance in several tasks (e.g., WinoGrand, ARC-E and Wikitext2). It is also notable that the performance drop of ChatGLM3-6B-base after pruning is relatively lower than that of the LLaMA series. One possible reason for this is that the ChatGLM models have a higher dimension in the MLP UP projection, which may result in more redundancy within the model and more robust to pruning.  

We also provide the results for a pruning rate of 30\% in Table~\ref{tab:vicuna30} and Table~\ref{tab:llama30} for the Vicuna-7B-v1.5 and LLaMA2-7B-chat-hf models, respectively. The proposed method outperform other methods in PIQA and WinoGrande tasks. The perplexity on the wikitext2 dataset is also relatively low. It is notable that the bias compensation term does not always introduce performance gain. For a better visualization of the impact of the pruning rate, we show how the perplexity metric changes with the pruning rate in Figure~\ref{fig:ppl}. We can see that the impact of pruning rate on perplexity is smallest for our method and the FLAP method, followed by LLM-Pruner, then SliceGPT, with Shortened LLaMA being affected the most.

\subsection{Impact of Hyperparameters}

The number of independent components (n\_components) in ICA analysis is the only hyperparameter in the proposed method. We empirically set n\_components as 128 in above experiments. To investigate the most suitable setting for the n\_components parameter, we conduct a simple analysis on a small sample of calibration data. We use the neural signals obtained from 40 input samples on the Vicuna-7B-v1.5 model and decompose these signals into 10, 20, 64, 128, 256, and 512 functional networks, respectively. Then, model pruning is performed based on these functional networks, and the pruned models are evaluated in terms of perplexity on the Wikitext2 dataset. The results are shown in Table~\ref{tab:vicuna_ncomponents}.

\begin{table}
    \centering
    \label{tab:vicuna_ncomponents}
    \begin{tabular}{ccc}
        \toprule
        Pruning Rate & n\_components & Perplexity \\
        \midrule
        20\% & 10 & 22.2733 \\
        20\% & 20 & 21.6782 \\
        20\% & 64 & 23.2159 \\
        20\% & 128 & 18.9421 \\
        20\% & 256 & 18.4713 \\
        20\% & 512 & 20.2174 \\
        \bottomrule 
    \end{tabular}
\end{table}

It is seen that decomposing the signals into 128 or 256 functional networks yields the best performance. However, when increasing the number of independent components to 512, the performance decreases. This could potentially due to the unconvergence of CanICA. As the number of independent components increases, the CanICA model becomes less likely to converge. Based on multiple experiments, we set the number of independent components in CanICA as 128 in this study. 


\subsection{Impact of Number of Calibration Samples}
\begin{figure}
    \centering
    \includegraphics[width=\linewidth]{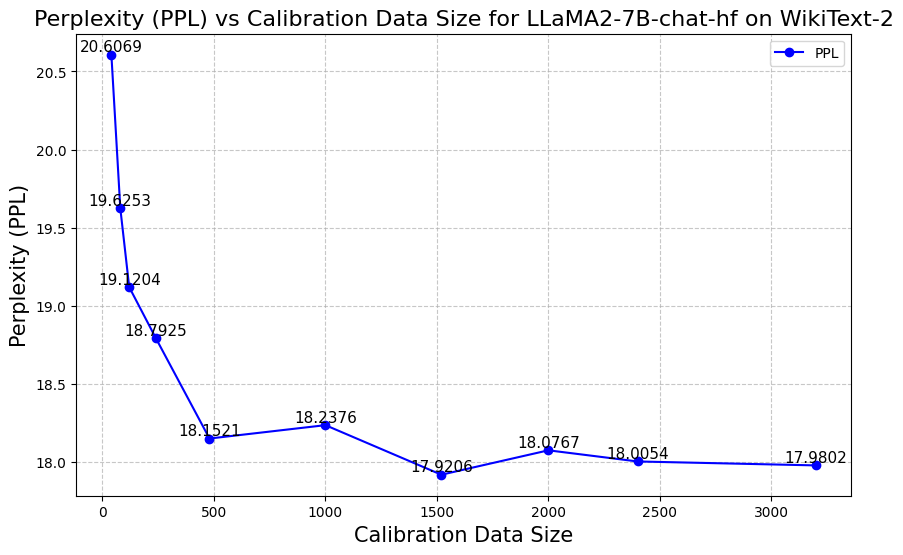}
    \caption{The impact of the number of calibration samples on pruning performance. The x-axis represents the number of calibration samples, while the y-axis shows the perplexity results tested on the WikiText-2 dataset. The model used for this experiment is LLaMA2-7B-chat-hf. Pruning rate is 20\%.}
    \label{fig:calibration_size}
\end{figure}
Figure~\ref{fig:calibration_size} presents the perplexity results of LLaMA2-7B-chat-hf on the WikiText-2 dataset after pruning with varying sizes of calibration datasets. Specifically, the model was pruned using different numbers of calibration samples, ranging from 40 to 3200, and the resulting perplexity was measured to evaluate the performance of the pruned model. We set n\_components as 128 and pruning rate as 20\% in this experiment. 

We can observe that, although there is a general trend indicating that the model's perplexity tends to decrease as the amount of data increases, the perplexity does not necessarily reach its minimum value with a larger number of calibration samples. Instead, the lowest perplexity occurs at around 1500 samples. 

Additionally, we also observe that, as shown in Table~\ref{tab:llama20}, FLAP achieved a final perplexity of 18.2824 using 1024 calibration samples for model pruning. However, our method surpasses this performance with fewer samples, achieving a perplexity of 18.1521 at just 480 calibration samples. This indicates that our approach can achieve better pruning results with a smaller number of calibration samples.

\section{Discussion and Conclusion}

Inspired by functional brain network studies in neuroscience, we propose a structured pruning method for LLMs by identifying and preserving their functional networks in this study. We treat an LLM as a digital brain and apply functional brain network analytic framework to identify functional networks in LLMs. By preserving the key neurons in those functional networks, we can effectively prune LLMs. Our experimental results show that the proposed method outperforms state-of-the-art structured pruning methods in a set of commonly used tasks at different pruning rates.

The proposed method faces some limitations. First, the idea of identifying and preserving functional networks for LLM pruning is realized in a local form, that is, we prune LLMs layer-by-layer for computational cost considerations. However, the proposed idea also supports a global strategy by identifying and preserving functional networks in neurons aggregated from all layers in LLMs. Considering the information flow in Transformer block, it is reasonable to hypothesize that there exist functional networks across layers in LLMs. Thus, global identifying and preserving functional networks in LLMs are expected to better maintain functional coherence compared to the local strategy applied in this study and may further decrease performance loss in model pruning. Secondly, it would be of great interest to assess the behaviors of the functional networks identified in LLMs, by applying a similar approach that is adopted in explaining neurons in smaller language models (e.g., GPT-2) using larger models (e.g., GPT-4) \cite{bills2023language}. In this context, the method proposed in this study may also provide a framework for mechanistic interpretation of LLMs. Thirdly, while the ICA algorithm adopted in study has been widely used in exploring functional brain networks in fMRI data, it is only a linear method and may not be adequate to capture functional networks depicting non-linear dependencies among massive neurons in LLMs. Meanwhile, non-linear deep neural networks, such as auto-encoders \cite{liu2023spatial,liu2024spatial,liu2024mapping,he2023multi,qiang2022learning,qiang2024deep}, have been successfully applied for functional brain network analysis. Thus, it would also be interest to adopt such non-linear methods to identify functional networks in LLMs for the purpose of pruning. 

Last but not least, our research draws inspiration from neuroscience studies on functional brain networks. There are numerous similarities between the human brain and LLMs. Methods used for analyzing the human brain may also be applicable to analyzing LLMs, and the same applies in reverse. In this sense, the analytical methods for both LLMs and the human brain can mutually inspire and enhance each other, promoting their co-development. In the future, it might be possible to introduce more principles of the human brain or techniques used in brain research into AI studies, which could hold significant implications for the field of artificial intelligence research.

\bibliography{aaai2026}

\end{document}